\documentclass[letterpaper, 10 pt, conference]{support/ieeeconf}
\IEEEoverridecommandlockouts
\usepackage{cite}
\usepackage{amsmath,amssymb,amsfonts}
\usepackage{graphicx}
\usepackage{textcomp}
\usepackage{xcolor}
\usepackage{multirow}
\usepackage{etoolbox}
\usepackage{url}
\usepackage{algorithm}
\usepackage[noend]{algorithmic}
\usepackage{pifont}
\usepackage{graphicx}
\usepackage[colorlinks,
            linkcolor=black,
            anchorcolor=black,
            urlcolor=black,
            citecolor=black]{hyperref}

\makeatletter
\makeatletter
\def\endthebibliography{%
	\def\@noitemerr{\@latex@warning{Empty `thebibliography' environment}}%
	\endlist
}

\patchcmd{\@makecaption}
  {\scshape}
  {}
  {}
  {}
\makeatletter
\patchcmd{\@makecaption}
  {\\}
  {.\ }
  {}
  {}
\makeatother
\def\tablename{Table}

\def\BibTeX{{\rm B\kern-.05em{\sc i\kern-.025em b}\kern-.08em
    T\kern-.1667em\lower.7ex\hbox{E}\kern-.125emX}}
\begin{document}
\setlength{\skip\footins}{3pt} 

\title{Efficient Trajectory Generation Based on Traversable Planes \\ in 3D Complex Architectural Spaces}
\author{Mengke Zhang\textsuperscript{1,2,$\dagger$}, Zhihao Tian\textsuperscript{2,$\dagger$}, Yaoguang Xia\textsuperscript{3}, Chao Xu\textsuperscript{1,2}, Fei Gao\textsuperscript{1,2}, Yanjun Cao\textsuperscript{1,2}
        \thanks{$^*$This work was supported by National Natural Science Foundation of China under Grant 62103368. }
        \thanks{$^1$The State Key Laboratory of Industrial Control Technology, College of Control Science and Engineering, Zhejiang University, Hangzhou 310027, China.}
        \thanks{$^2$Huzhou Institute, Zhejiang University, and Huzhou Key Laboratory of Autonomous System, Huzhou 313000, China. }
        \thanks{$^3$China Tobacco Zhejiang Industrial Co., Ltd., Hangzhou 310024, China.}
        \thanks{$^\dagger$ Equal contribution.}
        \thanks{Email:\tt\fontsize{7.8pt}{10pt}\selectfont\{mkzhang233, yanjunhi\}@zju.edu.cn}
        \thanks{\;\;\;\;\;\;\;\;\;\; \tt\fontsize{7.8pt}{10pt}\selectfont zhihaotian37@gmail.com}
}

\maketitle

\begin{abstract}
	
With the increasing integration of robots into human life, their role in architectural spaces where people spend most of their time has become more prominent.
While motion capabilities and accurate localization for automated robots have rapidly developed, the challenge remains to generate efficient, smooth, comprehensive, and high-quality trajectories in these areas.
In this paper, we propose a novel efficient planner for ground robots to autonomously navigate in large complex multi-layered architectural spaces.
Considering that traversable regions typically include ground, slopes, and stairs, which are planar or nearly planar structures, we simplify the problem to navigation within and between complex intersecting planes.  
We first extract traversable planes from 3D point clouds through segmenting, merging, classifying, and connecting to build a plane-graph, which is lightweight but fully represents the traversable regions. 
We then build a trajectory optimization based on motion state trajectory and fully consider special constraints when crossing multi-layer planes to maximize the robot's maneuverability.
We conduct experiments in simulated environments and test on a CubeTrack robot in real-world scenarios, validating the method's effectiveness and practicality.
\end{abstract}


\vspace{-0.25cm}

\section{Introduction}

Robots are desired to assist with various aspects of human life, particularly in architectural spaces where people spend the majority of their time, such as houses, factories, offices, shopping malls, and complex buildings.
Recently, advancements in motion control for tracked robots have significantly improved their ability to navigate obstacles like stairs and slopes.
Additionally, the development of visual and LiDAR-based SLAM technology offers stable and accurate localization information.
To apply these robots as fully autonomous systems in large 3D complex multi-layered architectural spaces, how to generate efficient, smooth, comprehensive, and high-quality trajectories connecting any two positions in the space remains unresolved.


\begin{figure}[h]
  \centering
  \setlength{\abovecaptionskip}{-0.1cm}
  \includegraphics[width=8cm]{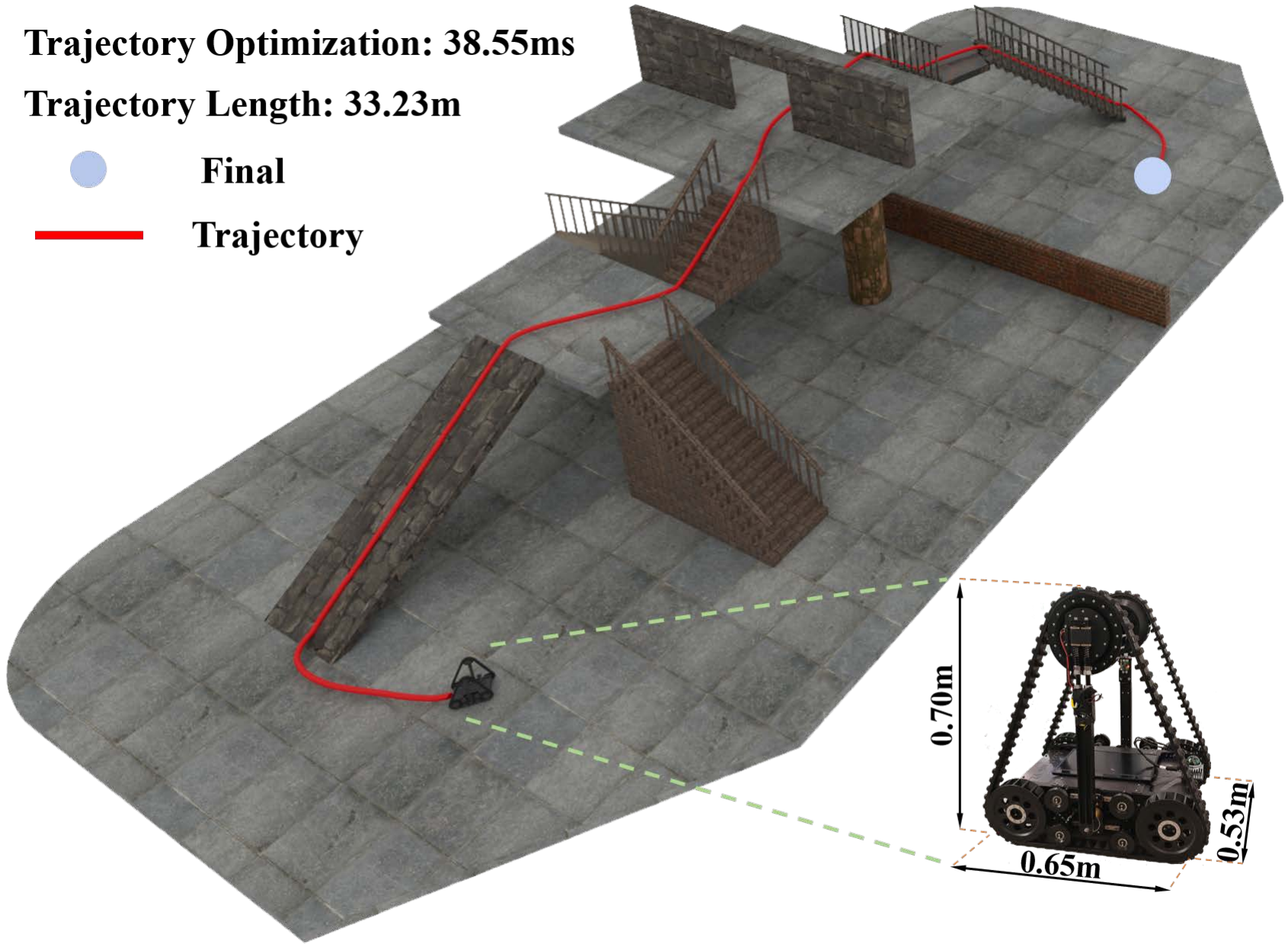}
  \caption{
    The proposed trajectory generation enables the robot to navigate in a complex environment, passing through high platforms via stairs and slopes to avoid an obstacle wall and finally arrive at the target in the map. 
    }
  \label{fig:HeadFigure}
  \vspace{-0.75cm}
\end{figure}

Several methods have proposed planning techniques for ground robots in 3D spaces, but primarily focused on single-layer settings.
A common approach is to directly plan on point clouds or meshes; however, these methods often consume excessive memory and have limitations in accurately representing ground features. 
Given that ground robots are restricted to ground movement, generating trajectories directly on point clouds or meshes with optimal postures is difficult.
Elevation maps can efficiently represent terrain information and are used to generate the trajectories well.
However, it only works in single-layer environments, such as roads or single floors.
To adapt the approach for multi-layer environments, additional rule-based layering strategies are necessary.


We target an efficient planner for ground robots to generate a smooth, unified, continuous trajectory in multi-layered spaces without breaking down the process in one task.
Considering that traversable regions typically include roads, floors, slopes, and stairs, which are planar or nearly planar structures, we simplified the problem to navigation within and between complex intersecting planes.  
We first extract traversable planes from 3D point clouds through segmenting, merging, classifying, and connecting to build a plane graph, which is lightweight but fully represents the traversable regions. 
This allows us to transform the trajectory generation from the 3D spaces to intersecting planes, effectively reducing the complexity and accelerating the path search process. 
We introduce a trajectory representation that represents trajectories as positional increments within plane coordinate systems, which is used for trajectory optimization within multiple planes.
In addition, we fully consider special constraints when crossing multi-layer planes to maximize the robot's maneuverability and ensure safe operation on risky structures. In summary, our contributions are as follows:

1. We propose an efficient trajectory generation planner specifically for 3D complex architectural spaces, which leverages traversable planes of simplify the problem to navigation within and between planes.

2. We design an approach to extract a traversable plane graph from point clouds, simplifying the representation of feasible regions in 3D spaces. 

3. We propose a plane graph path search approach and an optimization-based trajectory generation method, efficiently generating safe and kinematically appropriate trajectories.

4. We validate our method on various complex simulation maps and verified its practicality in the real environment.

The rest of the paper is organized as follows.
In Sec.\ref{sec:Related_Work}, we review related work.
In Sec.\ref{sec:Extract_Traversable_Planes}, we introduce the process of extracting traversable planes and the construction of plane graph.
In Sec.\ref{sec:Trajectory_Generation}, we introduce the proposed trajectory generation, including path searching and trajectory optimization.
Sec.\ref{sec:Experiments} and Sec.\ref{sec:Conclusion} present the experiments and conclusions.

\section{Related Work}\label{sec:Related_Work}

In recent years, planning methods for ground robots in 3D environments have been extensively studied. 

The global point clouds are easily obtained in most architectural 3D spaces. 
Tensor voting \cite{liu2015robotic, colas20133d} is used to estimate the robot's pose. 
Chen \cite{chen2023geometry} determine the robot's possible poses from point clouds, which are used for planning and control.
Although planning directly on the point cloud can yield accurate attitude information, it is time-consuming. 
These methods find it challenging to generate continuous trajectories within a limited time frame.

Traversable planes are widely used for the planning of ground robots.
Jian \cite{jian2022putn} fit planes on point clouds and generate a traversable strip for guiding trajectory generation.
Learning based approaches \cite{hoeller2022neural, wen2022robust} are used to predict traversable areas and robot poses. 
However, these methods simply consider the single-layer 3D environment with different elevations, instead of architectural spaces with multiple layers. 
Road segmentation \cite{deng2022hd, kim2024make, deng2024opengraph} is also widely used, but is also difficult to extend to multi-layer 3D complex architectural spaces.

Several methods \cite{rosmann2013efficient, kurenkov2022nfomp, han2023efficient, zhang2024universaltrajectoryoptimizationframework} can generate smooth and high-quality trajectories for ground robots.
However they assume horizontal planes in the optimization, making them unsuitable for 3D environments with slopes.
Distance fields\cite{kurenkov2022nfomp, zhang2024universaltrajectoryoptimizationframework} are used to ensure the robot's safety by maintaining a minimum distance from obstacles. 
Based on the concept of fields, some methods \cite{atas2022elevation, wang2023towards, xu2023efficient, leininger2024gaussian} discretize the environment and estimate the pose to construct fields for trajectory generation. 
Yang \cite{yang2024efficient} use tomographic analysis to divide 3D structures into multiple layers, reducing the map size and accelerating path search.
However, these methods usually lose structured information because of the discretization. 
The motion constraints for robots on specific planes may differ from those on a horizontal plane, which may result in generating risky trajectories.

Indoor reconstruction techniques can effectively extract traversable planes in architectural spaces. 
Industry Foundation Classes models (IFC)\cite{diakite2016extraction} and Feature Structure Map (FSM)\cite{shi2020novel} are used to reconstruct 3D free spaces for navigation.
Jelena \cite{gregoric2024autonomous} construct a hierarchical graph based on known planes for path planning. 
However, these methods are limited to extracting floor planes and overly simplify the modeling of stairs. 
To determine the navigation between layers, stair detection\cite{westfechtel2018robust, sriganesh2023fast, lee2022vision, kim2024staircase} is proposed to calculate the position of stairs and is used to find the optimal routes in complex buildings\cite{nikoohemat2020indoor}.


\section{Traversable Plane Graph Construction} \label{sec:Extract_Traversable_Planes}
In this section, we introduce extracting traversable planes from point clouds. 
As shown in Alg. \ref{alg:extractSurface}, we take global point clouds as input and output traversable planes and their parameters, including the transformation matrix $\boldsymbol{T_p}$ from the world coordinate ststem $\Psi_w$ to the plane coordinate system $\Psi_p$, the Euclidean Signed Distance Field (ESDF), the indices and connecting lines of adjacent planes.
In the following text, we use $^*\boldsymbol{p}$ to indicate the coordinate system $\Psi_{*}$ to which the point belongs. 
\vspace{-0.3cm}
\begin{algorithm}[h]
  \small
  \caption{Extract Traversable Planes}
  \begin{algorithmic}
      \STATE \hspace{-0.4cm} \text{\textbf{Input}: global point clouds: $cloud$}
      \STATE \hspace{-0.4cm} \text{\textbf{Output}: traversable planes: }
      \STATE $PL_t=\{(\boldsymbol{T_p}, ESDF, index, line)\}$
  \end{algorithmic}
  \begin{algorithmic}[1]
      \STATE $normalmap \gets \textbf{PreAnalyse}(cloud)$
      \STATE $PL \gets \textbf{RegionGrowing}(cloud, normalmap)$

      \STATE $PL_t, PL_v \gets \textbf{Merge}(PL)$

      \FOR{each $P_i \in PL_t$}
          \STATE $P_i.boundary \gets \textbf{CalConvex}(P_i)$
      \ENDFOR
      \FOR{each $P_i,P_j \in PL_t$}
          \IF{$\textbf{Adjacent}(P_i,P_j)$}
              \STATE $P_i.line, P_j.line \gets \textbf{CalConnect}(P_i, P_j)$
              \STATE $P_i.index \gets j, P_j.index \gets i$
          \ENDIF
      \ENDFOR  
      \FOR{$P_i \in PL_t$}
          \STATE $\boldsymbol{T_s} \gets \textbf{SetGridmap}(P_i)$
          \FOR{$j \in P_i.index$}
              \STATE $\textbf{SetOverlap}(P_i, PL_t[j])$
          \ENDFOR
          \STATE $\textbf{SetOccupied}(P_i, PL_v)$
          \STATE $P_i.ESDF \gets \textbf{UpdateESDF}(P_i)$
      \ENDFOR
  \end{algorithmic}
  \label{alg:extractSurface}
\end{algorithm}
\vspace{-0.35cm}

\begin{figure}[b]
  \centering
  \vspace{-0.6cm}
  \setlength{\abovecaptionskip}{-0.1cm}
  \includegraphics[width=8.5cm]{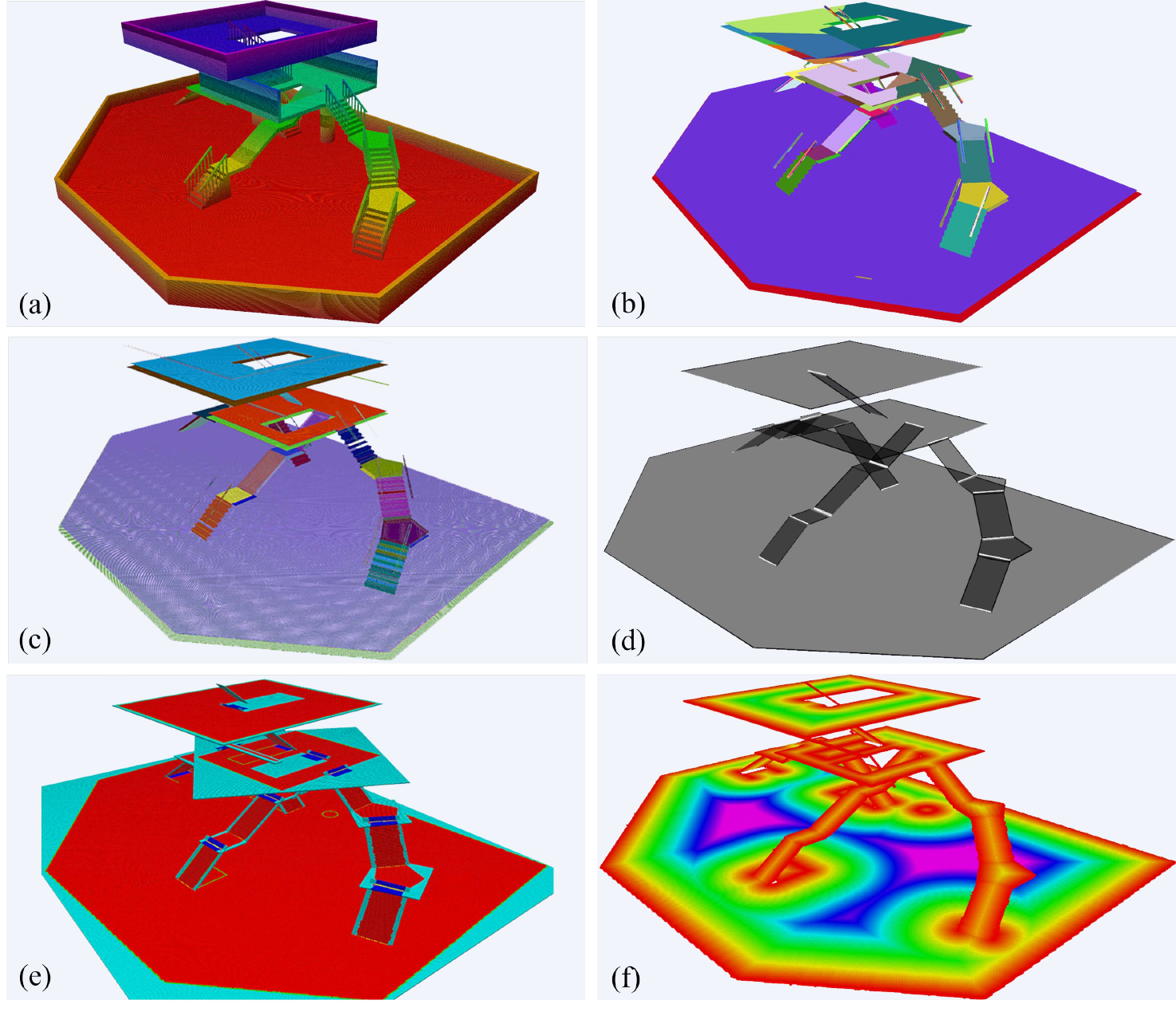}
  \caption{The process of extracting traversable planes. 
  (a) Original point clouds. 
  (b) Planes extracted using the region-growing. 
  (c) Merged traversable planes. 
  (d) Connectivity between traversable planes, where white lines are intersection lines. 
  (e) Gridding. 
  (f) ESDF.}
  \label{fig:extract_planes}
\end{figure}

\textbf{Plane Extraction and Merging} (line 1 to 3): 
We first downsample and filter the input point clouds, and then estimate normals to get the normal map (Fig.\ref{fig:extract_planes}(a)). 
With the region-growing\cite{westfechtel2018robust}, we extract planes \(PL\) from the point clouds and the normal map (Fig.\ref{fig:extract_planes}(b)). 
Inspired by \cite{westfechtel2018robust}, we merge the planes belonging to stairs. 
The planes are classified into traversable planes \(PL_t\) and vertical planes \(PL_v\) with the inclination calculated by covariance matrixes. 
We merge coplanar and adjacent \(PL_t\) to get the complete plane structure. 
If two parallel planes of the same size are close, we remove the lower plane considering thickness (Fig.\ref{fig:extract_planes}(c)).

\textbf{Determining Plane Connectivity} (line 4 to 9):
We project points of each \( PL_t \) onto the plane and calculate a minimal convex polygon containing these points.
Because of possible misclassification of points near the boundary, we slightly expand the boundary points of the polygon outward.
If an intersection \( \boldsymbol{l}^{i,j} \) of the two polygon is found, we save the two endpoints \( {^w\boldsymbol{l}_1^{i,j}}, {^w\boldsymbol{l}_2^{i,j}} \) and the indices of the intersecting planes (Fig.\ref{fig:extract_planes}(d)). 

\begin{figure}[t]
  \centering
  \vspace{0.2cm}
  \setlength{\abovecaptionskip}{-0.1cm}
  \includegraphics[width=8.5cm]{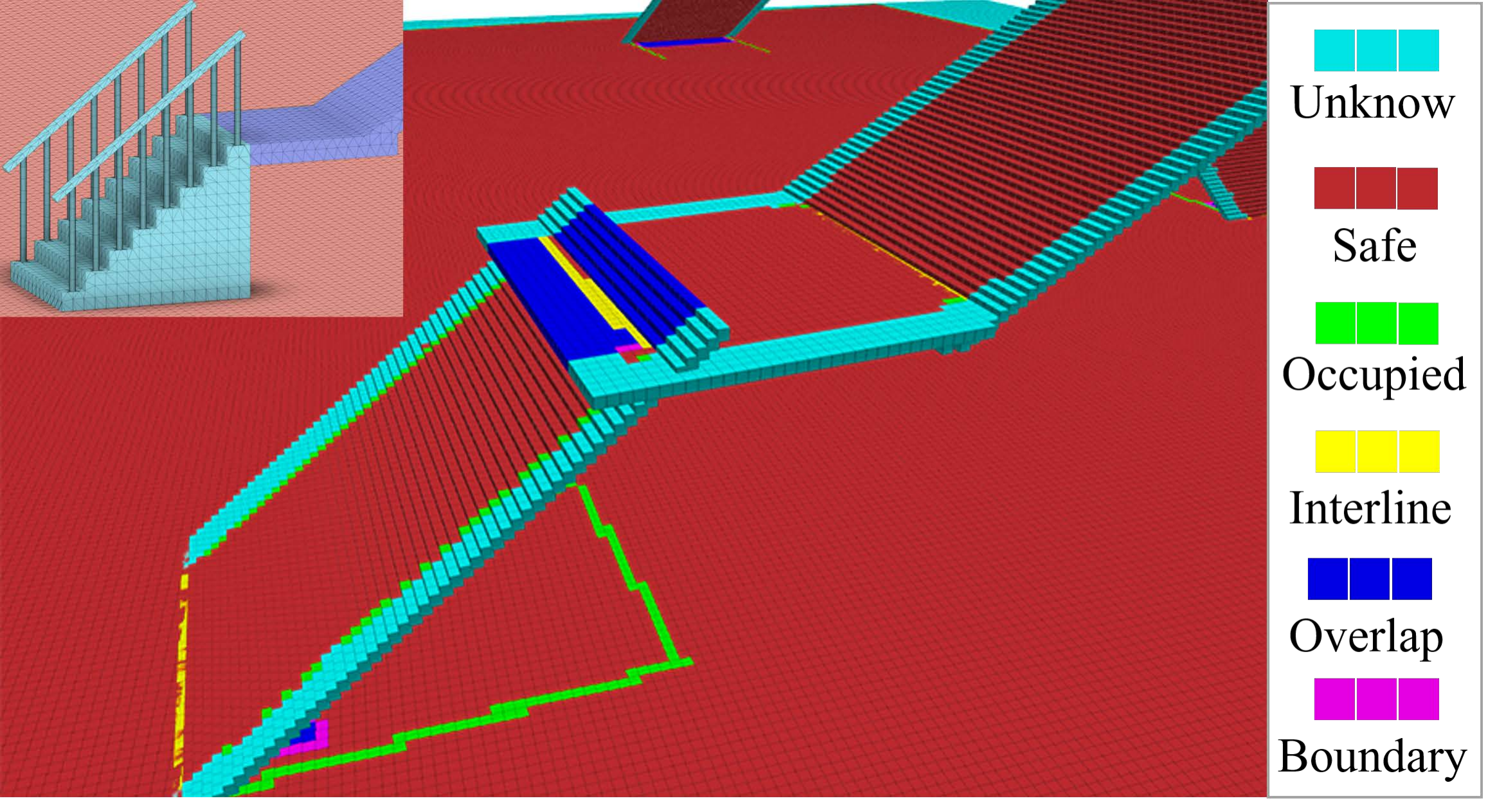}
  \caption{Grid states. 
  Considering the expanded boundaries, the grid map is larger than the plane itself. 
  The top left corner shows the model.}
  \label{fig:gridState}
  \vspace{-0.75cm}
\end{figure}

\textbf{Gridding}(line 10 to 15):
We use the centroid of the point clouds \( c_i \in \mathbb{R}^3 \) as the origin of \( \Psi_i \), the plane's ascent direction as the x-axis and the direction perpendicular to the plane as the z-axis, by which we can get the rotation matrix \( \boldsymbol{R_i}\in\mathbb{R}^{3\times 3} \) and translation vector \( \boldsymbol{c_i}\in\mathbb{R}^3 \) of the transformation matrix \( \boldsymbol{T_i} \). 
The grid map is initialized as $Unknown$. 
The grid will be set as $Safe$ if there is at last one projection point in it. 
The grid corresponding to the intersection lines between planes are marked as $Interline$, while overlapping non-traversable regions are marked as $Overlap$. 
The $Overlap$ grids adjacent to $Safe$ grids are defined as $Boundary$. 
Considering that vertical planes \( PL_v \) may be above \( PL_t \) as obstacles, we use alpha shapes \cite{edelsbrunner1994three} to find the boundary points of \( PL_v \).
If the distance from the boundary points to \( PL_t \) is less than the safety distance, the corresponding grid will be set as $Occupied$ (Fig.\ref{fig:extract_planes}(e)). 
Examples of grid states are shown in Fig.\ref{fig:gridState}.
Finally, we update ESDF by considering $Unknown$, $Occupied$, and $Boundary$ grids as obstacles (Fig.\ref{fig:extract_planes}(f)).

So far, we represent the feasible regions of the 3D environment as traversable planes, which preserve the feasible regions as simple a structure as possible and as little data as possible. 

We use an undirected cost graph \( G=(V,E,W) \) to describe these planes.
We define the vertex set \( V \) as points on the intersection lines \( \boldsymbol{l}^{i,j} \). 
Initially, these points are assumed to be at the midpoint of \( \boldsymbol{l}^{i,j} \). 
We use ESDF to check if the point is feasible in both adjacent planes; if it is not feasible, we check points on either side until the first feasible point is found. 
After getting all \( V \), we traverse all the planes and use A* to determine if a path exists connecting the two vertices within the plane, and record the path as an edge \( E \), the cost is corresponding weight \( W \).

\section{Trajectory Generation}\label{sec:Trajectory_Generation}
In this section, we introduce how to use plane graph for path searching and generate cross-plane trajectories that satisfy constraints through optimization.

\subsection{Path Searching}\label{sec:Path_Searching}

After getting the starting position \( {^w\boldsymbol{p}_0} \), we first project the point onto the nearest plane \( P_1 \) and use its projection point \( ^1\boldsymbol{p}_0 \) as the start node. 
We search for paths from \( ^1\boldsymbol{p}_0 \) to all vertices on $P_1$. 
The feasible paths and their costs are added to $G$. 
The final position \( {^w\boldsymbol{p}_f} \) is similarly processed as the final node. 
With breadth-first search within \( G \), we connect edges to find the path from the start node to the final node.

\subsection{Trajectory representation}\label{sec:Trajectory_representation}

As differential drive robots exhibit superior maneuverability in narrow environments, we use the MS trajectory \cite{zhang2024universaltrajectoryoptimizationframework}, which is suitable for these robots.
Intuitively, trajectories on a plane represent the incremental position starting from the start position on the plane. 
The MS trajectory is a polynomial function of the yaw angle $^w\theta$ in $\Psi_w$ and forward arc length $s$ relative to time. 
The $mi$-th segment of the trajectory can be represented as: 
\begin{align}
  {^w\theta_{mi}(t)} = \boldsymbol{\beta}^T(t) \boldsymbol{c}_{\theta, mi}, \\
  s_{mi}(t) = \boldsymbol{\beta}^T(t) \boldsymbol{c}_{s, mi},
\end{align}
where \(\boldsymbol{c}_{mi} = [\boldsymbol{c}_{\theta, mi}, \boldsymbol{c}_{s, mi}]\) is the coefficients of the polynomial, and \(\boldsymbol{\beta}(t)\) is the natural basis. 
For simplicity, we use \(\boldsymbol{\sigma} = [^w\theta, s]^T\). 
The high-order continuity of the MS trajectory is inherently satisfied, and a smooth bijection is used for the non-negativity constraints of trajectory duration.

To represent the trajectory across multiple planes, we partition the whole trajectory into \(\mathcal T\) parts, with the \(ti\)-th part representing the trajectory on the \(ti\)-th plane.
Each trajectory part consists of \(M_{ti}\) polynomial segments, so the entire trajectory comprises \(M = \sum_{ti=1}^{\mathcal T} M_{ti}\) segments. 
Each polynomial segment is confined to a single plane. 
Unlike \cite{zhang2024universaltrajectoryoptimizationframework}, due to each plane having its own coordinate system, the trajectory must be located on the plane. 
Therefore, the trajectory should be transformed into the local coordinate system for planning.

Within the \(ti\)-th plane, the trajectory should start from the point \( {^{w}\boldsymbol{p}^{ti}_{0}}\) on the intersection line or starting point \( {^{w}\boldsymbol{p}_{0}}\), with its projection on the plane given by \({^{ti}}\boldsymbol{p}_0 = \{ {^{ti}}x_0, {^{ti}}y_0 \}\). 
For simplicity, \( {^{w}\boldsymbol{p}_{0}}\) and \( {^{w}\boldsymbol{p}^{1}_{0}}\) are equivalent next.
The trajectory in $\Psi_{ti}$ can be expressed as:
\begin{align}
{^{ti}}x(t) = \int_0^t [\dot{s}_{ti}(\tau) \cos({^w\theta_{ti}(\tau)} - \Delta\theta_{ti})] d\tau + {^{ti}}x_0, \\
{^{ti}}y(t) = \int_0^t [\dot{s}_{ti}(\tau) \sin({^w\theta_{ti}(\tau)} - \Delta\theta_{ti})] d\tau + {^{ti}}y_0,
\end{align}
where \(\Delta\theta_{ti}\) is the yaw angle offset of \(\Psi_{ti}\) relative to \(\Psi_w\). 
We use Simpson's rule to approximate \(\{{^{ti}}x(t), {^{ti}}y(t)\}\), allowing the robot's state in $\Psi_{ti}$ to be calculated at any time.

\subsection{Trajectory Optimization}
Based on the trajectory representation proposed in Sec.\ref{sec:Trajectory_representation}, we set the objective function $\mathcal J$ as:
\begin{align}
\mathop{min}\limits_{\boldsymbol{c}, \boldsymbol{T}} \mathcal{J} = \int_0^{T_s} \boldsymbol{\sigma}^{(3)}(t)^T \boldsymbol{W} \boldsymbol{\sigma}^{(3)}(t)dt + \epsilon_T T_s  + w_d \mathcal{C}_d,
\end{align}
where $\boldsymbol W\in\mathbb R^{2\times 2}$ is a diagonal matrix to penalize control efforts, $T_s=\sum_{mi=1}^{M} T_{mi}$ is the trajectory duration and $\epsilon_T$ is the weight.
The objective is to minimize the jerk of \(\boldsymbol{\sigma}\) to reduce control effort while introducing a time term to balance the trajectory duration. 
We employ the penalty function method, where we ensure \(\mathcal{C}_d(\boldsymbol{\sigma}(t), \ldots, \ddot{\boldsymbol{\sigma}}(t)) \leq 0\) by including constraint \(d\) into the optimization function, with \(w_d\) representing the corresponding weight.

In cross-plane planning, the trajectory starts from ${^{w}}\boldsymbol{p}^{ti}_0$ and ends at the global final position or point on the intersection line ${^{w}}\boldsymbol{p}^{ti}_f$, where ${^{w}}\boldsymbol{p}^{ti+1}_0 = {^{w}}\boldsymbol{p}^{ti}_f$. 
Considering that ${^{w}}\boldsymbol{p}^{ti}_f$ on the intersection line is got from searching rather than optimization.
To ensure the trajectory's optimality, ${^{w}}\boldsymbol{p}^{ti}_f$ should be parts of the optimization variables. 
Assuming the endpoints of the intersection line are \({^w\boldsymbol{l}_0^{ti}}, {^w\boldsymbol{l}_1^{ti}}\), we introduce a parameter \(\eta_{ti}\) as:
\begin{equation}
  \begin{aligned}[c]
    {^{w}}\boldsymbol{p}^{ti}_f & = {^w\boldsymbol{l}_0^{ti}} + \eta^*_{ti} ({^w\boldsymbol{l}_1^{ti}} - {^w\boldsymbol{l}_0^{ti}}), \eta^*_{ti} \in (0,1), \\
    \eta^*_{ti} & = \frac{1}{1 + e^{-\eta_{ti}}},  \eta_{ti} \in \mathbb{R},  ti \in \{1, \ldots, \mathcal T-1\},
  \end{aligned}
\end{equation}
where \(\eta^*_{ti}\) is the proportion of the position from \({^w\boldsymbol{l}_0^{ti}}\) to \({^w\boldsymbol{l}_1^{ti}}\). 
Since $\eta^*_{ti} \in (0,1)$, we introduce an unconstrained variable \(\eta_{ti}\) as the optimization variable to simplify the problem.

As the MS trajectory requires integration to compute positions, we introduce ALM \cite{rockafellar1974augmented} to ensure that trajectories on the plane reach the given final position. 
Suppose the final position calculated by integration in the $ti$-th plane is \( {^{ti}\tilde{\boldsymbol{p}}_f^{ti}} = [{^{ti}\tilde{x}_f^{ti}}, {^{ti}\tilde{y}_f^{ti}}]^T \). 
The desired final position is ${^{ti}}\boldsymbol{p}^{ti}_f  = [{^{ti}{x}_f^{ti}}, {^{ti}{y}_f^{ti}}]^T$ for \(ti \in \{1, \ldots, \mathcal T-1\}\). 
In the last plane, the desired final position is the global final position, ${^{\mathcal T}}\boldsymbol{p}^{\mathcal T}_f$.
In the following, we collectively refer to ${^{ti}}\boldsymbol{p}^{ti}_f$ for $ti \in \{1, \ldots, \mathcal T\}$.
For the x-axis of the \(ti\)-th plane as an example, the final position constraint can be expressed as:
\begin{align}
\mathcal{C}_{fx}^{ti}(\boldsymbol{c}, \boldsymbol{T}) = {^{ti}\tilde{x}_f^{ti}}(\boldsymbol{c}, \boldsymbol{T}) - {^{ti} x_f^{ti}}. 
\end{align}

Thus, the new optimization problem can be formulated as:
\begin{align}
\mathcal{J}_\rho(\boldsymbol{c}, \boldsymbol{T}, \boldsymbol{\eta}, \boldsymbol{\lambda}) = \mathcal{J} + \sum_{\substack{ti=1 \\ \iota=x,y}}^{\mathcal T} \frac{\rho}{2} \left\| \mathcal{C}_{f\iota}^{ti}(\boldsymbol{c}, \boldsymbol{T}) + \frac{\lambda_{\iota}^{ti}}{\rho} \right\|^2,
\label{ali:ALM}
\end{align}
where \(\boldsymbol{\lambda} = \{\lambda_\iota^{ti}\}\) are dual variables and \(\rho > 0\) is the weight of the augmentation term. 
We can solve the optimization problem Eq.(\ref{ali:ALM}) and update \(\boldsymbol{\lambda}\) to iteratively get the solution. 
For gradient propagation and solutions to the optimization problem, readers can refer to \cite{zhang2024universaltrajectoryoptimizationframework}. 
We set the convergence condition such that the error between ${^{ti}\tilde{\boldsymbol{p}}_f^{ti}}$ and ${^{ti}{\boldsymbol{p}}_f^{ti}}$ is less than the given value \(e_{max}\):
\begin{align}
\sqrt{({^{ti}\tilde{x}_f^{ti}} - {^{ti} x_f^{ti}})^2 + ({^{ti}\tilde{y}_f^{ti}} - {^{ti} y_f^{ti}})^2} < e_{max}. 
\end{align}
In practice, we generally set \(e_{max}=1\text{cm}\), ensuring that final position errors have minimal impact on subsequent control.

\subsection{Constraints}
In this section, we focus on the specific form of inequality constraints for navigation crossing multiple layers.

\textbf{Velocity Constraint}: Considering the impact of gravity, the robot's velocity on slopes varies with orientation. 
To describe this variation, we represent the maximum velocity on the slope as a function of the robot's yaw angle \(^p\theta\) in the plane coordinate system $\Psi_p$. 
For instance, when the robot is moving forward:
\begin{align}
  & v_{max}(^p\theta) =    \label{ali:vmax}\\
  & \begin{cases}
    v_{max} \sqrt{r_r(\psi)\cos^2{^p\theta} + \sin^2{^p\theta}}, & {^p\theta} \in [-\pi/2, \pi/2], \\
    v_{max} \sqrt{r_d(\psi)\cos^2{^p\theta} + \sin^2{^p\theta}}, & \text{otherwise},  \nonumber
  \end{cases}
\end{align}
where \(v_{max}\) is the maximum forward velocity on the horizontal ground, and \(\psi\) is the inclination angle of the slope. 
$r_r(\psi), r_d(\psi)$ are the ratio function when rising and declining along the plane, which is related to the specific robot. 
Eq.(\ref{ali:vmax}) is used to approximate the maximum velocity with two half-ellipses, ensuring that $v_{max}(^p\theta)$ is continuous and differentiable with respect to \(^p\theta\).

\begin{figure*}[h]
  \centering
  \vspace{0.2cm}
  \setlength{\abovecaptionskip}{-0.25cm}
  \includegraphics[width=16cm]{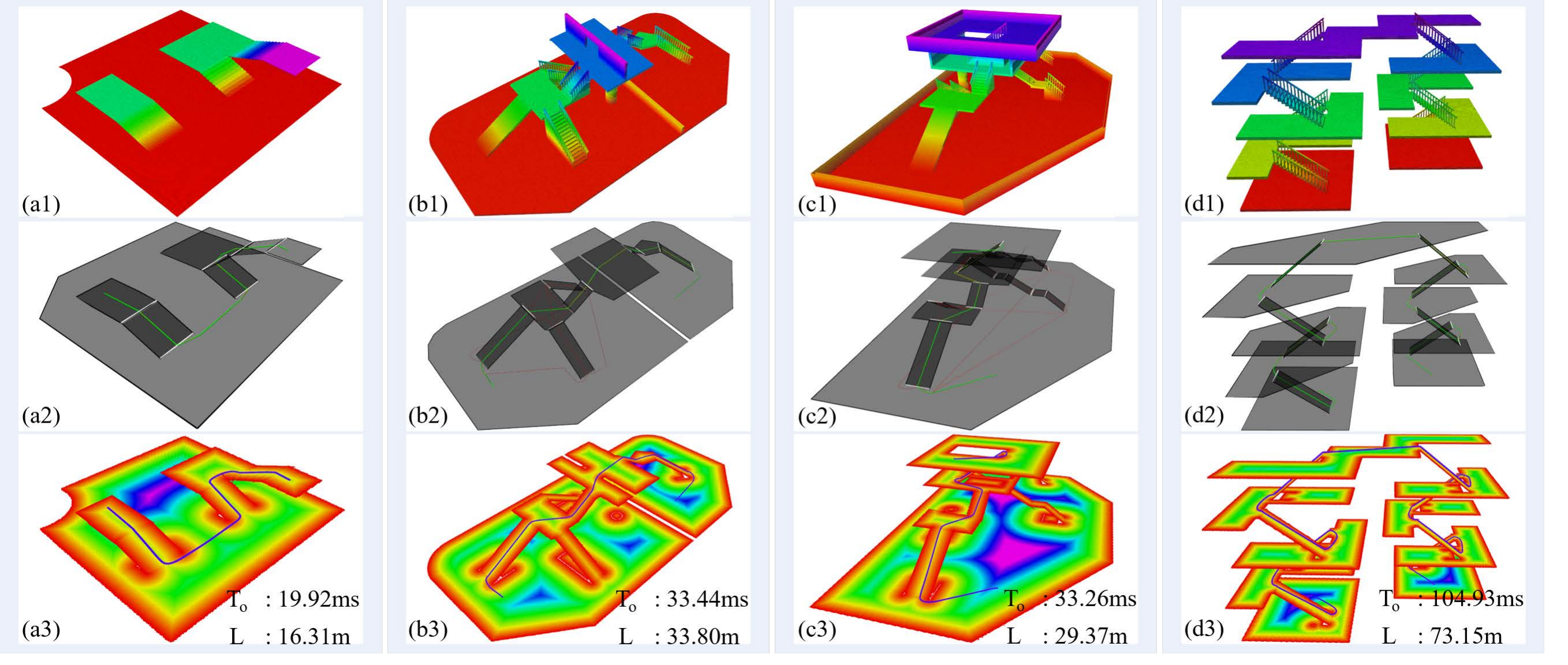}
  \caption{Simulation environment and trajectory generation of the proposed method. 
  The first row shows the point clouds of the environment: Planes(a1), Platform(b1), Multi-layer(c1), and Building(d1). 
  The second row shows traversable planes, where white lines are intersection lines between planes, red lines are the plane graph, and green lines are the searched path.
  The third row shows ESDF and the generated trajectory. 
  The table shows the trajectory length $L$ and trajectory optimization time $T_o$. 
  The proposed method is capable of generating feasible and safe trajectories in complex architectural spaces.}
  \label{fig:simulation}
  \vspace{-0.65cm}
\end{figure*}

Similar to \cite{zhang2024universaltrajectoryoptimizationframework}, considering that differential drive robots require differential wheel speeds to achieve rotation, the maximum angular velocity should depend on the current velocity. 
We can give the corresponding constraint as:
\begin{align}
  \mathcal{C}_{m+}(\boldsymbol{\sigma}, \boldsymbol{\dot\sigma}) = \varkappa\omega v_{max}(^p\theta) + \omega_M v - v_{max}(^p\theta) \omega_M, 
  \label{ali:moment}
\end{align}
where \(\varkappa \in \{-1, 1\}\), \(\omega_M\) is the maximum angular velocity when rotating in place, $\dot{\boldsymbol{\sigma}}=[\omega,v]$ is angular velocity and linear velocity. 
The constraint $\mathcal{C}_{m-}$ when the robot moves backward can also be obtained from Eq.(\ref{ali:moment}).

\textbf{Orientation Constraint}: When the robot is moving on stairs, the contact points are significantly fewer than those on a plane. 
If the orientation $^p\theta$ deviates too much from the upward direction of the stairs, it could result in side-slipping or even tipping over.
To ensure safety, we need to constrain the angle between them:
\begin{align}
  \mathcal{C}_O(\boldsymbol\sigma) = {^p\theta}'^2-{\theta_s}^2, {^p\theta} = {^p\theta}' + k\pi, {^p\theta}' \in [-\frac{\pi}{2}, \frac{\pi}{2}),  
  \label{ali:orientation}
\end{align}
where $k$ is an integer, and $\theta_s$ is the desired maximum angle. 
Due to the upward direction of the stairs is the x-axis of $\Psi_p$, Eq.(\ref{ali:orientation}) constrains the yaw angle in $\Psi_p$ to close to 0 or $\pi$. 

\textbf{Safety Constraint}: We use the ESDF from Sec.\ref{sec:Extract_Traversable_Planes} to ensure safety.
The constraint requires that the ESDF value of the current position $\{{^px}, {^py}\}$ in $\Psi_p$ is greater than a safety distance \(d_s\): 
\begin{align}
  \mathcal{C}_s({^px}, {^py}) = d_s - E_p({^px}, {^py}), 
  \label{ali:safety}
\end{align}
where \(E_p({^px}, {^py})\) is the ESDF value obtained through bilinear interpolation in plane $p$. 


\section{Experiments}\label{sec:Experiments}
We conduct extensive experiments in both simulated and real-world scenarios to validate the proposed method. 
We test the proposed method in complex 3D environments and compare our approach with state-of-the-art trajectory generation methods in 3D environments to verify its effectiveness. 
Finally, we deploy our method on a tracked robot in real-world scenarios to prove its practicality.

\subsection{Simulation}

\begin{figure}[b]
  \centering
  \vspace{-0.5cm}
  \setlength{\abovecaptionskip}{-0.1cm}
  \includegraphics[width=8.5cm]{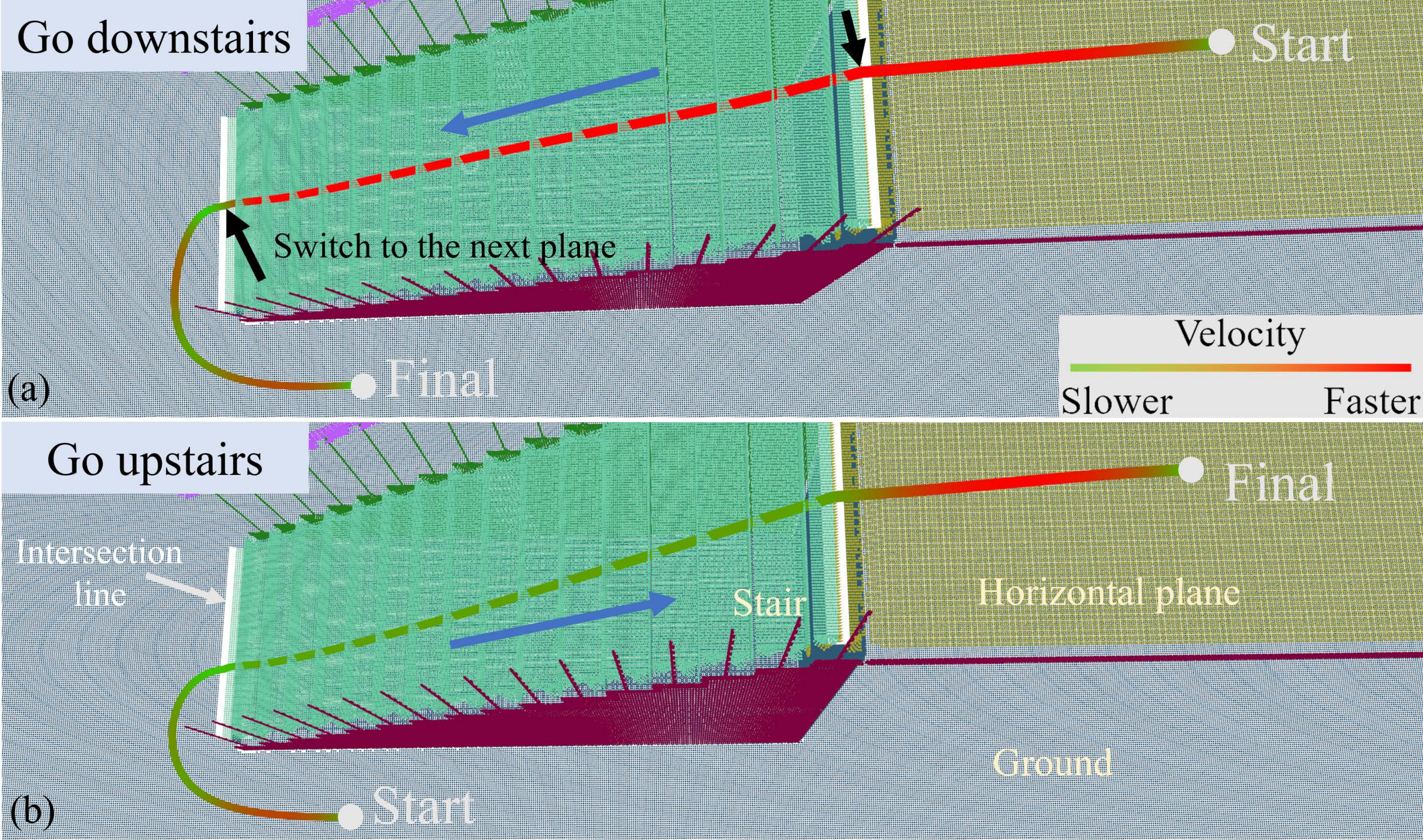}
  \caption{
    The trajectory of going upstairs and downstairs. 
    The trajectory has a higher velocity when moving straight on the horizontal plane or going downstairs compared to the lower velocity when going upstairs.
    The trajectory switches to the next plane at the intersection line, ensuring velocity continuity during the switch. 
  }
  \label{fig:velocity}
\end{figure}

To demonstrate the effectiveness of the proposed trajectory generation, we construct four complex 3D structured environments to verify the proposed method:

\begin{itemize} 
  \item A two-layered plane with noise(Planes): A simple indoor scene comprising a ramp and two stairs. We add noise to simulate the actual point cloud.
  \item A platform with obstacles(Platform): A two-story structure where the robot should traverse from one side of the wall to the other via a platform.
  \item Multi-layered plane(Multi-layer): A multi-layered structure with complex routes, requiring the robot to choose the correct path among ramps and stairs to reach a higher level. 
  \item Building: A common structure with stairs, requiring the robot to use the stairs to move between floors.
\end{itemize}

The extracted traversable planes and generated trajectories in different environments are shown in Fig.\ref{fig:simulation}.

As shown in Fig. \ref{fig:velocity}, the application of ALM enables the connection of trajectories between planes with minimal error, while the continuity of the polynomial trajectory ensures smooth higher-order kinematics.
We demonstrate the trajectory and velocity profile when going upstairs and downstairs. 
The proposed method effectively limits the velocity on inclined planes within a reasonable range based on the inclination angle of the plane.

\begin{figure}[b]
  \centering
  \vspace{-0.7cm}
  \setlength{\abovecaptionskip}{-0.15cm}
  \includegraphics[width=8cm]{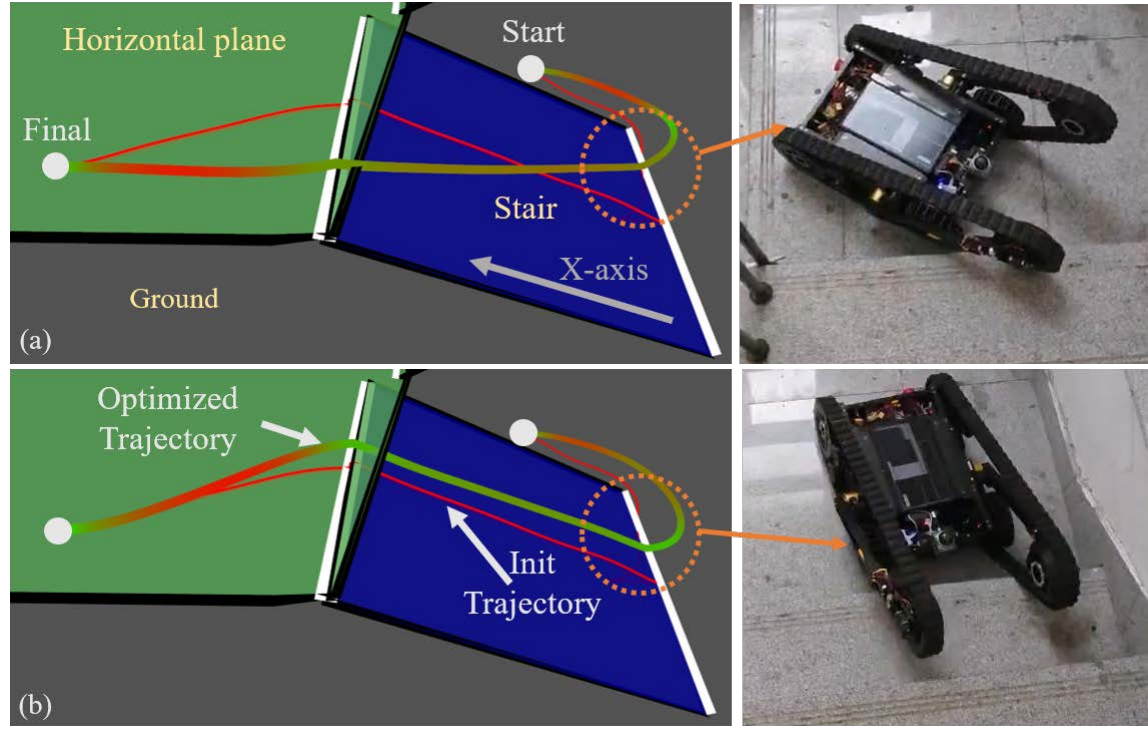}
  \caption{
    Comparison of trajectories without (a) and with (b) orientation constraints.
    The red line is the initial trajectory from the searched path. 
    Orientation constraints can effectively limit the robot's orientation on stairs, ensuring its safety.
  }
  \label{fig:orientation}
\end{figure}

Thanks to the characteristics of traversable planes, the proposed method can constrain the robot's orientation to ensure safety on risky planes. 
As shown in Fig.\ref{fig:orientation}(a), without the orientation constraint in Eq.(\ref{ali:orientation}), the robot may approach the stairs at a large angle, risking hindrance or tipping over. 
By constraining the orientation, the robot can efficiently climb the stairs, as illustrated in Fig.\ref{fig:orientation}(b).

\subsection{Benchmarks}
In this section, we compare the proposed method with Wang's\cite{wang2023towards}, Xu's\cite{xu2023efficient}, and Yang's\cite{yang2024efficient}. 
Wang constructs a penalty field within a 3D environment to optimize the trajectory in \( \mathbb{R}^3 \). 
Xu calculates the terrain of the elevation map and generates trajectories that satisfy nonholonomic dynamics with ALM. 
Yang slices the 3D structure into multiple layers with a tomographic method and plans on these layers. 
A comparison of the characteristics of the four methods is shown in Table \ref{tab:Qualitative_comparison}. 

\setlength{\tabcolsep}{3pt} 
\renewcommand{\arraystretch}{0.8} 
\vspace{-0.2cm}
\begin{table}[h]
  \scriptsize
  \caption{Comparison of trajectory generation characteristics. }
  \vspace{-0.2cm}
  \begin{tabular}{c|ccccc}
  \hline
           & \begin{tabular}[c]{@{}c@{}}Planning \\ in \\ 3D space\end{tabular} & \begin{tabular}[c]{@{}c@{}}Velocity limit \\ for \\ inclined planes\end{tabular} & \begin{tabular}[c]{@{}c@{}}Planning \\ on \\ the stairs\end{tabular} & \begin{tabular}[c]{@{}c@{}}Reasonable \\ orientation\\ angle constraint\end{tabular} & \begin{tabular}[c]{@{}c@{}}Constrained \\ angular \\ velocity\end{tabular} \\[1ex] \hline 
  Porposed &  \ding{51}  &  \ding{51}  &  \ding{51}  &  \ding{51}  &  \ding{51}  \\[1ex]
  Wang's\cite{wang2023towards}   &  \ding{51}  &  \ding{55}  &   \ding{55}   &  \ding{55}   &  \ding{51}  \\[1ex]
  Xu's\cite{xu2023efficient}     &  \ding{55}  &  \ding{51}  &   \ding{51}  &  \ding{55}  &  \ding{51}  \\[1ex]
  Yang's\cite{yang2024efficient}   &  \ding{51}  &  \ding{55}  &  \ding{51}  &  \ding{55}  &  \ding{55}  \\ \hline
  \end{tabular}
  \label{tab:Qualitative_comparison}
  \vspace{-0.4cm}
\end{table}

\begin{figure}[t]
  \centering
  \vspace{0.2cm}
  \setlength{\abovecaptionskip}{-0.1cm}
  \includegraphics[width=8.5cm]{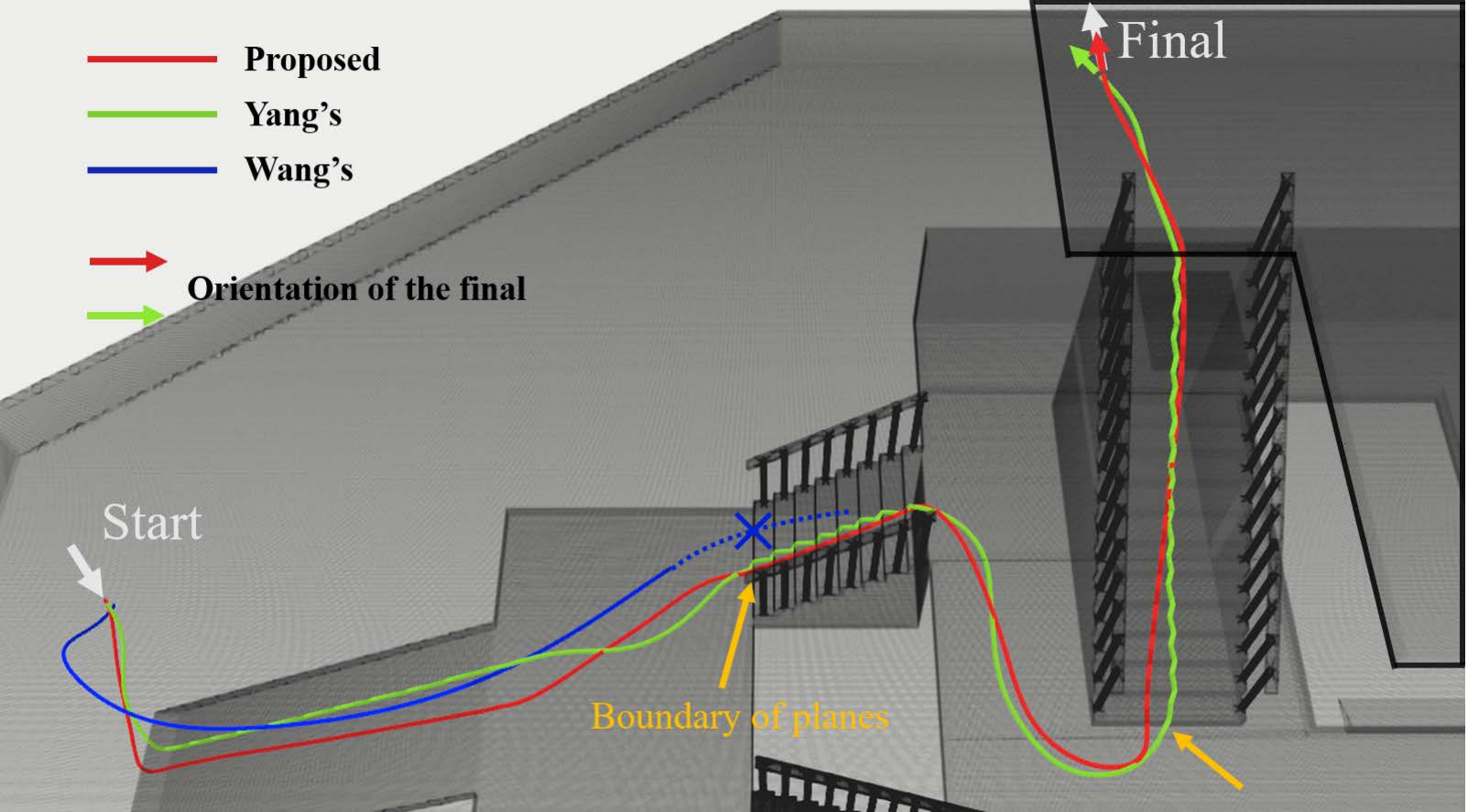}
  \caption{ Comparison of three methods.  
  The proposed method generates smoother trajectories and aligns with the inclined plane in advance to ensure safety. 
  Wang's method can traverse slopes but not stairs. }
  \label{fig:benchmark}
  \vspace{-0.6cm}
\end{figure}

To validate the effectiveness of our method, we conduct tests in the environment shown in Fig.\ref{fig:simulation}(c).
We selected the starting position at the lowest layer and set the goal at the highest layer. 
The results of several  methods that can plan in 3D environments are shown in Fig.\ref{fig:benchmark} and Table \ref{tab:Num_comparison}. 
By simplifying the planes as a plane graph, the proposed method demonstrates a significant advantage in path searching.
For trajectory optimization, the proposed method generates smoother trajectories within a similar computation time. 
The proposed trajectory effectively constrains angular velocity and the orientation of the start and end points, unlike Yang's method, which only plans positions, as shown in the final of Fig.\ref{fig:benchmark}.
For fairness, we convert all data to 32-bit floating point when calculating map size. 
For Yang's method, we omit ceiling data as it is unnecessary and gradient data as it can be computed from other data.
Yang's method requires storing height data and may have duplicates, while our method represents the map as nearly non-overlapping traversable planes with boundaries and sizes, requiring a smaller size to store the map. 

\renewcommand{\arraystretch}{1.0} 
\vspace{-0.2cm}
\begin{table}[h]
  \caption{Comparison of trajectory generation characteristics. }
  \vspace{-0.2cm}
  \begin{tabular}{c|cccc}
  \hline
           & \begin{tabular}[c]{@{}c@{}}Trajectory\\ length/m\end{tabular} & \begin{tabular}[c]{@{}c@{}}Time of path \\ finding/ms\end{tabular} & \begin{tabular}[c]{@{}c@{}}Time of trajectory \\ optimization/ms\end{tabular} & \begin{tabular}[c]{@{}c@{}}Map size\\ /MB\end{tabular} \\ \hline
  Proposed &        25.68                 & 2.07                                             & 38.83                                        & 1.7                                                  \\
  Yang's\cite{yang2024efficient}  &        27.39          & 59.72                                         & 49.78                                         & 6.0                                                  \\ \hline
  \end{tabular}
  \label{tab:Num_comparison}
  \vspace{-0.4cm}
\end{table}


\subsection{Real-World Experiment}
\begin{figure}[t]
  \centering
  \vspace{0.2cm}
  \setlength{\abovecaptionskip}{-0.15cm}
  \includegraphics[width=8.5cm]{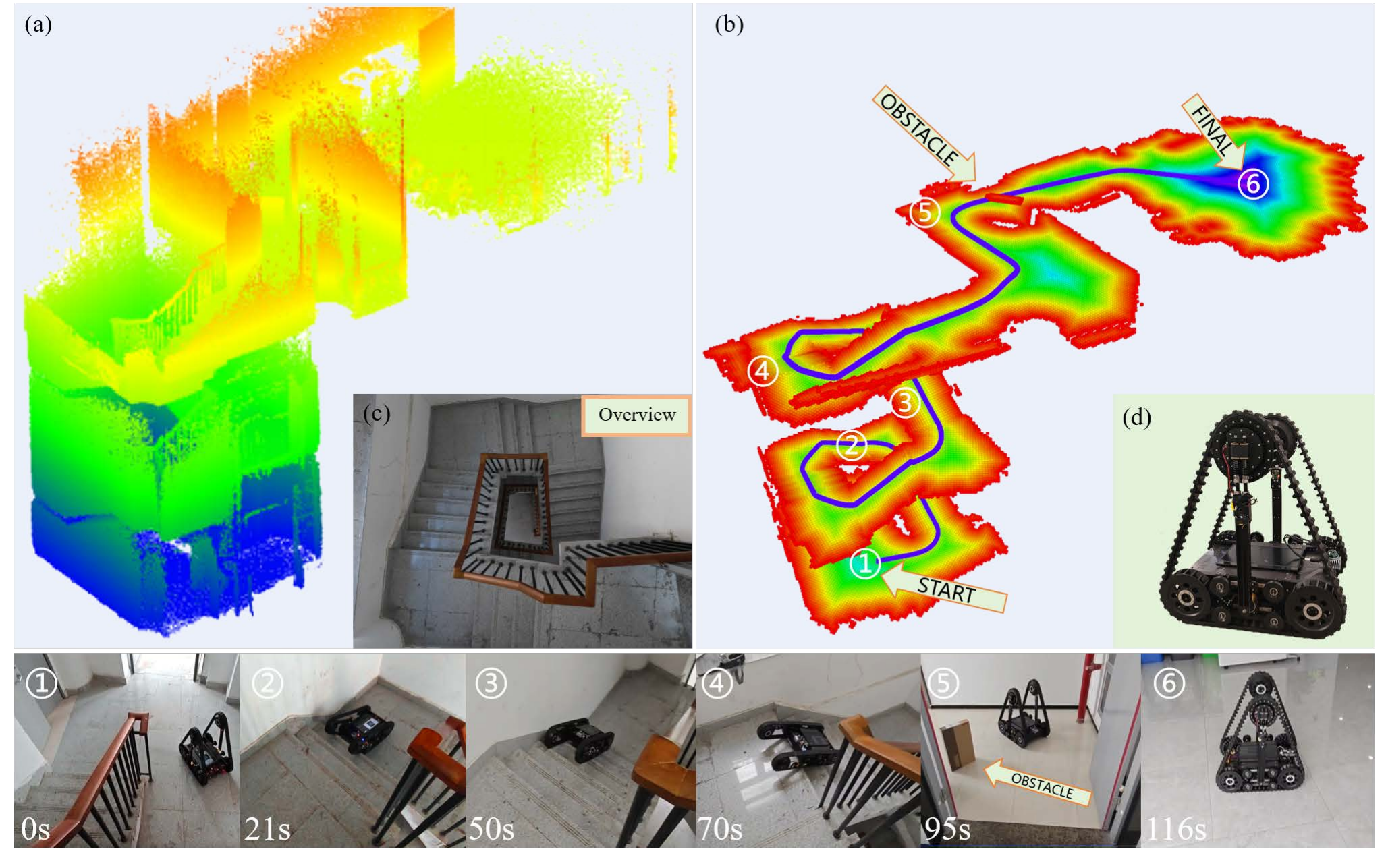}
  \caption{Results of the real-world experiment. 
  (a)The original point clouds. 
  (b)ESDF is calculate based on traversable planes and the generated trajectory. 
  (c)The complex non-Manhattan stair comprises nine consecutive stair segments with handrails connected by platforms. 
  (d) The CubeTrack robot was used for the experiment. 
  \ding{172}-\ding{176} are snapshots of the motion process.
  }
  \label{fig:real_exam}
  \vspace{-0.65cm}
\end{figure}

We conduct the real-world experiment using CubeTrack \cite{cubetrack}, a tracked robot equipped with variable geometry tracks, as shown in Fig.\ref{fig:real_exam}(d). 
The orientation of the flippers is controlled by the motor located at the end of the flippers. 
FAST-LIO2\cite{Xu2022FASTLIO2} is used for localization. 
All codes are executed on the NUC13.

The robot runs in a complex non-Manhattan environment, as illustrated in Fig.\ref{fig:real_exam}, which comprises nine consecutive stair segments with handrails, connected by platforms to form a spiral structure ascending to the second floor. 
There are two narrow doorways on the second floor, one of which is obstructed by the obstacle before reaching the final position. 

Based on the point clouds scanned from the real environment, as shown in Fig.\ref{fig:real_exam}(a)(c), we generated trajectories within the complex 3D environment constituted by the extracted traversable planes, as shown in Fig.\ref{fig:real_exam}(b). 
The trajectory length is 30.11m, and the required trajectory optimization time is 51.24 ms.
After generating the trajectory, we can determine the robot's position at given times and predict its subsequent posture based on the plane it belongs to.
The role of flippers is considered to smooth the robot's posture changes when switching traversable planes, so the predicted posture is used to estimate the flipper angle, as shown in Fig.\ref{fig:real_exam}\ding{173}-\ding{175}. 


\section{Conclusion}\label{sec:Conclusion}

In this paper, we present an efficient trajectory generation method based on traversable planes for robots navigating in complex architectural spaces. 
We extract traversable planes from point clouds and build the plane graph to simplify the problem of navigation in 3D environments. 
We introduce cross-plane path searching and trajectory generation and design corresponding trajectory optimization problems and constraints to generate safe and efficient trajectories. 
Experiments in both simulated and real-world scenarios demonstrate that the proposed method can generate feasible trajectories. 
Future work will focus on more architectural spaces, such as spiral ramps and rotating staircases. 

\bibliography{references}

\begin{thebibliography}{10}
\providecommand{\url}[1]{#1}
\csname url@rmstyle\endcsname
\providecommand{\newblock}{\relax}
\providecommand{\bibinfo}[2]{#2}
\providecommand\BIBentrySTDinterwordspacing{\spaceskip=0pt\relax}
\providecommand\BIBentryALTinterwordstretchfactor{4}
\providecommand\BIBentryALTinterwordspacing{\spaceskip=\fontdimen2\font plus
\BIBentryALTinterwordstretchfactor\fontdimen3\font minus \fontdimen4\font\relax}
\providecommand\BIBforeignlanguage[2]{{%
\expandafter\ifx\csname l@#1\endcsname\relax
\typeout{** WARNING: IEEEtran.bst: No hyphenation pattern has been}%
\typeout{** loaded for the language `#1'. Using the pattern for}%
\typeout{** the default language instead.}%
\else
\language=\csname l@#1\endcsname
\fi
#2}}

\bibitem{liu2015robotic}
M.~Liu, ``Robotic online path planning on point cloud,'' \emph{IEEE transactions on cybernetics}, vol.~46, no.~5, pp. 1217--1228, 2015.

\bibitem{colas20133d}
F.~Colas, S.~Mahesh, F.~Pomerleau, M.~Liu, and R.~Siegwart, ``3d path planning and execution for search and rescue ground robots,'' in \emph{2013 IEEE/RSJ International Conference on Intelligent Robots and Systems}.\hskip 1em plus 0.5em minus 0.4em\relax IEEE, 2013, pp. 722--727.

\bibitem{chen2023geometry}
B.~Chen, K.~Huang, H.~Pan, H.~Ren, X.~Chen, J.~Xiao, W.~Wu, and H.~Lu, ``Geometry-based flipper motion planning for articulated tracked robots traversing rough terrain in real-time,'' \emph{Journal of Field Robotics}, vol.~40, no.~8, pp. 2010--2029, 2023.

\bibitem{jian2022putn}
Z.~Jian, Z.~Lu, X.~Zhou, B.~Lan, A.~Xiao, X.~Wang, and B.~Liang, ``Putn: A plane-fitting based uneven terrain navigation framework,'' in \emph{2022 IEEE/RSJ International Conference on Intelligent Robots and Systems (IROS)}.\hskip 1em plus 0.5em minus 0.4em\relax IEEE, 2022, pp. 7160--7166.

\bibitem{hoeller2022neural}
D.~Hoeller, N.~Rudin, C.~Choy, A.~Anandkumar, and M.~Hutter, ``Neural scene representation for locomotion on structured terrain,'' \emph{IEEE Robotics and Automation Letters}, vol.~7, no.~4, pp. 8667--8674, 2022.

\bibitem{wen2022robust}
M.~Wen, Y.~Dai, T.~Chen, C.~Zhao, J.~Zhang, and D.~Wang, ``A robust sidewalk navigation method for mobile robots based on sparse semantic point cloud,'' in \emph{2022 IEEE/RSJ International Conference on Intelligent Robots and Systems (IROS)}.\hskip 1em plus 0.5em minus 0.4em\relax IEEE, 2022, pp. 7841--7846.

\bibitem{deng2022hd}
Y.~Deng, M.~Wang, Y.~Yang, and Y.~Yue, ``Hd-ccsom: Hierarchical and dense collaborative continuous semantic occupancy mapping through label diffusion,'' in \emph{2022 IEEE/RSJ International Conference on Intelligent Robots and Systems (IROS)}.\hskip 1em plus 0.5em minus 0.4em\relax IEEE, 2022, pp. 2417--2422.

\bibitem{kim2024make}
M.~Kim, S.~Lee, J.~Ha, and H.~Lee, ``Make your autonomous mobile robot on the sidewalk using the open-source lidar slam and autoware,'' \emph{IEEE Transactions on Intelligent Vehicles}, 2024.

\bibitem{deng2024opengraph}
Y.~Deng, J.~Wang, J.~Zhao, X.~Tian, G.~Chen, Y.~Yang, and Y.~Yue, ``Opengraph: Open-vocabulary hierarchical 3d graph representation in large-scale outdoor environments,'' \emph{arXiv preprint arXiv:2403.09412}, 2024.

\bibitem{rosmann2013efficient}
{R{\"o}smann, Christoph and Feiten, Wendelin and W{\"o}sch, Thomas and Hoffmann, Frank and Bertram, Torsten}, ``Efficient trajectory optimization using a sparse model,'' in \emph{2013 European Conference on Mobile Robots}.\hskip 1em plus 0.5em minus 0.4em\relax IEEE, 2013, pp. 138--143.

\bibitem{kurenkov2022nfomp}
M.~Kurenkov, A.~Potapov, A.~Savinykh, E.~Yudin, E.~Kruzhkov, P.~Karpyshev, and D.~Tsetserukou, ``Nfomp: Neural field for optimal motion planner of differential drive robots with nonholonomic constraints,'' \emph{IEEE Robotics and Automation Letters}, vol.~7, no.~4, pp. 10\,991--10\,998, 2022.

\bibitem{han2023efficient}
Z.~Han, Y.~Wu, T.~Li, L.~Zhang, L.~Pei, L.~Xu, C.~Li, C.~Ma, C.~Xu, S.~Shen, \emph{et~al.}, ``An efficient spatial-temporal trajectory planner for autonomous vehicles in unstructured environments,'' \emph{IEEE Transactions on Intelligent Transportation Systems}, 2023.

\bibitem{zhang2024universaltrajectoryoptimizationframework}
\BIBentryALTinterwordspacing
M.~Zhang, Z.~Han, C.~Xu, F.~Gao, and Y.~Cao, ``Universal trajectory optimization framework for differential-driven robot class,'' 2024. [Online]. Available: \url{https://arxiv.org/abs/2409.07924}
\BIBentrySTDinterwordspacing

\bibitem{atas2022elevation}
F.~Atas, G.~Cielniak, and L.~Grimstad, ``Elevation state-space: Surfel-based navigation in uneven environments for mobile robots,'' in \emph{2022 IEEE/RSJ International Conference on Intelligent Robots and Systems (IROS)}.\hskip 1em plus 0.5em minus 0.4em\relax IEEE, 2022, pp. 5715--5721.

\bibitem{wang2023towards}
J.~Wang, L.~Xu, H.~Fu, Z.~Meng, C.~Xu, Y.~Cao, X.~Lyu, and F.~Gao, ``Towards efficient trajectory generation for ground robots beyond 2d environment,'' in \emph{2023 IEEE International Conference on Robotics and Automation (ICRA)}.\hskip 1em plus 0.5em minus 0.4em\relax IEEE, 2023, pp. 7858--7864.

\bibitem{xu2023efficient}
L.~Xu, K.~Chai, Z.~Han, H.~Liu, C.~Xu, Y.~Cao, and F.~Gao, ``An efficient trajectory planner for car-like robots on uneven terrain,'' in \emph{2023 IEEE/RSJ International Conference on Intelligent Robots and Systems (IROS)}.\hskip 1em plus 0.5em minus 0.4em\relax IEEE, 2023, pp. 2853--2860.

\bibitem{leininger2024gaussian}
A.~Leininger, M.~Ali, H.~Jardali, and L.~Liu, ``Gaussian process-based traversability analysis for terrain mapless navigation,'' \emph{arXiv preprint arXiv:2403.19010}, 2024.

\bibitem{yang2024efficient}
B.~Yang, J.~Cheng, B.~Xue, J.~Jiao, and M.~Liu, ``Efficient global navigational planning in 3-d structures based on point cloud tomography,'' \emph{IEEE/ASME Transactions on Mechatronics}, 2024.

\bibitem{diakite2016extraction}
A.~A. Diakit{\'e} and S.~Zlatanova, ``Extraction of the 3d free space from building models for indoor navigation,'' \emph{ISPRS Annals of the Photogrammetry, Remote Sensing and Spatial Information Sciences}, vol.~4, pp. 241--248, 2016.

\bibitem{shi2020novel}
P.~Shi, Q.~Ye, and L.~Zeng, ``A novel indoor structure extraction based on dense point cloud,'' \emph{ISPRS International Journal of Geo-Information}, vol.~9, no.~11, p. 660, 2020.

\bibitem{gregoric2024autonomous}
J.~Gregori{\'c}, M.~Seder, and I.~Petrovi{\'c}, ``Autonomous hierarchy creation for computationally feasible near-optimal path planning in large environments,'' \emph{Robotics and autonomous systems}, vol. 172, p. 104584, 2024.

\bibitem{westfechtel2018robust}
T.~Westfechtel, K.~Ohno, B.~Mertsching, R.~Hamada, D.~Nickchen, S.~Kojima, and S.~Tadokoro, ``Robust stairway-detection and localization method for mobile robots using a graph-based model and competing initializations,'' \emph{The International Journal of Robotics Research}, vol.~37, no.~12, pp. 1463--1483, 2018.

\bibitem{sriganesh2023fast}
P.~Sriganesh, N.~Bagree, B.~Vundurthy, and M.~Travers, ``Fast staircase detection and estimation using 3d point clouds with multi-detection merging for heterogeneous robots,'' in \emph{2023 IEEE International Conference on Robotics and Automation (ICRA)}.\hskip 1em plus 0.5em minus 0.4em\relax IEEE, 2023, pp. 9253--9259.

\bibitem{lee2022vision}
K.~Lee, V.~Kalyanram, C.~Zhengl, S.~Sane, and K.~Lee, ``Vision-based ascending staircase detection with interpretable classification model for stair climbing robots,'' in \emph{2022 International Conference on Robotics and Automation (ICRA)}.\hskip 1em plus 0.5em minus 0.4em\relax IEEE, 2022, pp. 6564--6570.

\bibitem{kim2024staircase}
J.~Kim, S.~Jung, S.-K. Kim, Y.~Kim, and A.-a. Agha-mohammadi, ``Staircase localization for autonomous exploration in urban environments,'' \emph{arXiv preprint arXiv:2403.17330}, 2024.

\bibitem{nikoohemat2020indoor}
S.~Nikoohemat, A.~A. Diakit{\'e}, S.~Zlatanova, and G.~Vosselman, ``Indoor 3d reconstruction from point clouds for optimal routing in complex buildings to support disaster management,'' \emph{Automation in construction}, vol. 113, p. 103109, 2020.

\bibitem{edelsbrunner1994three}
H.~Edelsbrunner and E.~P. M{\"u}cke, ``Three-dimensional alpha shapes,'' \emph{ACM Transactions On Graphics (TOG)}, vol.~13, no.~1, pp. 43--72, 1994.

\bibitem{rockafellar1974augmented}
R.~T. Rockafellar, ``Augmented lagrange multiplier functions and duality in nonconvex programming,'' \emph{SIAM Journal on Control}, vol.~12, no.~2, pp. 268--285, 1974.

\bibitem{cubetrack}
C.~Xuan, J.~Lu, Z.~Tian, J.~Li, M.~Zhang, H.~Xie, J.~Qiu, C.~Xu, and Y.~Cao, ``Novel design of reconfigurable tracked robot with geometry-changing tracks,'' in \emph{2024 IEEE/RSJ International Conference on Intelligent Robots and Systems (IROS)}, 2024, pp. 10\,953--10\,960.

\bibitem{Xu2022FASTLIO2}
W.~Xu, Y.~Cai, D.~He, J.~Lin, and F.~Zhang, ``Fast-lio2: Fast direct lidar-inertial odometry,'' \emph{IEEE Transactions on Robotics}, vol.~38, no.~4, pp. 2053--2073, 2022.

\end{thebibliography}

\end{document}